\documentclass[conference]{IEEEtran}
\IEEEoverridecommandlockouts
\usepackage{cite}
\usepackage{amsmath,amssymb,amsfonts}
\usepackage{algorithmic}
\usepackage{graphicx}
\usepackage{textcomp}
\usepackage{xcolor}
\usepackage{multirow}

\begin{document}
\title{Virtual Foundry Graphnet for Metal Sintering Deformation Prediction
\thanks{}
}

\author{
\IEEEauthorblockN{ Rachel Chen}
\IEEEauthorblockA{\textit{HP Inc.}\\
Palo Alto, CA 94304, USA \\
lei.chen1@hp.com}
\and

\IEEEauthorblockN{ Chuang Gan}
\IEEEauthorblockA{\textit{HP Inc.}\\
2727 JinKe Rd, Shanghai, 201203, China \\
chuang.gan@hp.com}
\and

\IEEEauthorblockN{Juheon Lee}
\IEEEauthorblockA{\textit{Meta Inc.}\\
Menlo Park, CA, USA \\
juheon.lee@hp.com}
\and

\IEEEauthorblockN{ Zijiang Yang}
\IEEEauthorblockA{\textit{HP Inc.}\\
2727 JinKe Rd, Shanghai,2727 JinKe Rd, China \\
zijiang.yang@hp.com}
\and


\IEEEauthorblockN{Mohammad Amin Nabian}
\IEEEauthorblockA{\textit{NVIDIA}\\
Santa Clara, CA 95051, USA \\
mnabian@nvidia.com}
\and

\IEEEauthorblockN{ Jun Zeng}
\IEEEauthorblockA{\textit{HP Inc.}\\
Palo Alto, CA 94304, USA \\
@hp.com}
\and
}

\maketitle

\begin{abstract}
Metal sintering is a necessary step for Metal Injection Molded parts and binder jet such as HP’s metal 3D printer (MetJet). The metal sintering process introduces large deformation varying from 25\% to 50\% depending on the green part porosity. The final part's geometrical accuracy and consistency remain the top challenge to manufacturing yield. This is due to: (1) green parts out of MetJet printer are much more porous than other technologies (e.g., MIM); our green parts after sintering could result in ~50\% volumetric shrinkage. (2) Such shrinkage is not isotropic depending on non-uniform stress built up during the sintering process, e.g., gravitational sag, gravitational slump, surface drag.  In this paper, we use a graph-based deep learning approach to predict the part deformation, which can speed up the deformation simulation substantially at the voxel level. Running a well-trained Metal sintering inferencing engine only takes a range of seconds to obtain the final sintering deformation value. The tested accuracy on example complex geometry achieves 0.7um mean deviation for a 63mm testing part, for a single sintering step (equivalent to 8.3 minutes physical sintering time), and a 0.3mm mean deviation for the complete sintering cycle (approximately 4 hrs physical sintering time).  
\end{abstract}

\begin{IEEEkeywords}
graph neural networks, physics-ML, additive manufacturing, digital twin, metal sintering, deep learning
\end{IEEEkeywords}

\section{Introduction}
HP is a world leader in additive manufacturing and has a portfolio of products covering both polymer and metals, one of which is the celebrated HP’s Metal Jet 3D printing system S100. HP’s Metal Jet S100 uses precision control of heat to trigger the phase transition of metal powder material to create metal parts of shape and strength as designed. HP is developing Digital Twin software to virtualize such complex material phase transitions during production such that the user can predict and then optimize both the design parameters and the process control parameters to improve part quality and manufacturing yield. 

Metal sintering is a necessary step for Metal Injection Molded parts and binder jet such as HP’s MetJet\cite{INC_2022}. The metal sintering process introduces large deformation varying from 25\% to 50\% \cite{article}\cite{15256} depending on the green part porosity. The many factors causing the deformation (e.g., visco-plasticity, sintering pressure, yield surface parameters, yield stress, gravitational sag) must be captured and applied for shape deformation prediction. The final part, geometrical accuracy, and consistency remain the top challenge to manufacturing yield. This is due to: (1) green parts out of MetJet printer are much more porous than other technologies (e.g., MIM); our green parts after sintering could result in ~50\% volumetric shrinkage. (2) Such shrinkage is not isotropic depending on non-uniform stress built up during the sintering process, e.g., gravitational sag (feature location disruption due to insufficient support), gravitational slump (downward flow of material changing part thickness), surface drag (distortion caused by friction between a part and supporting surface during the movement induced by part shrinkage), as shown in Fig. \ref{fig2} the Stanford dragon test sample(4)\cite{The-stanford-3D-scanning-repository}.

\begin{figure}
\centering
\includegraphics[width=0.5\textwidth]{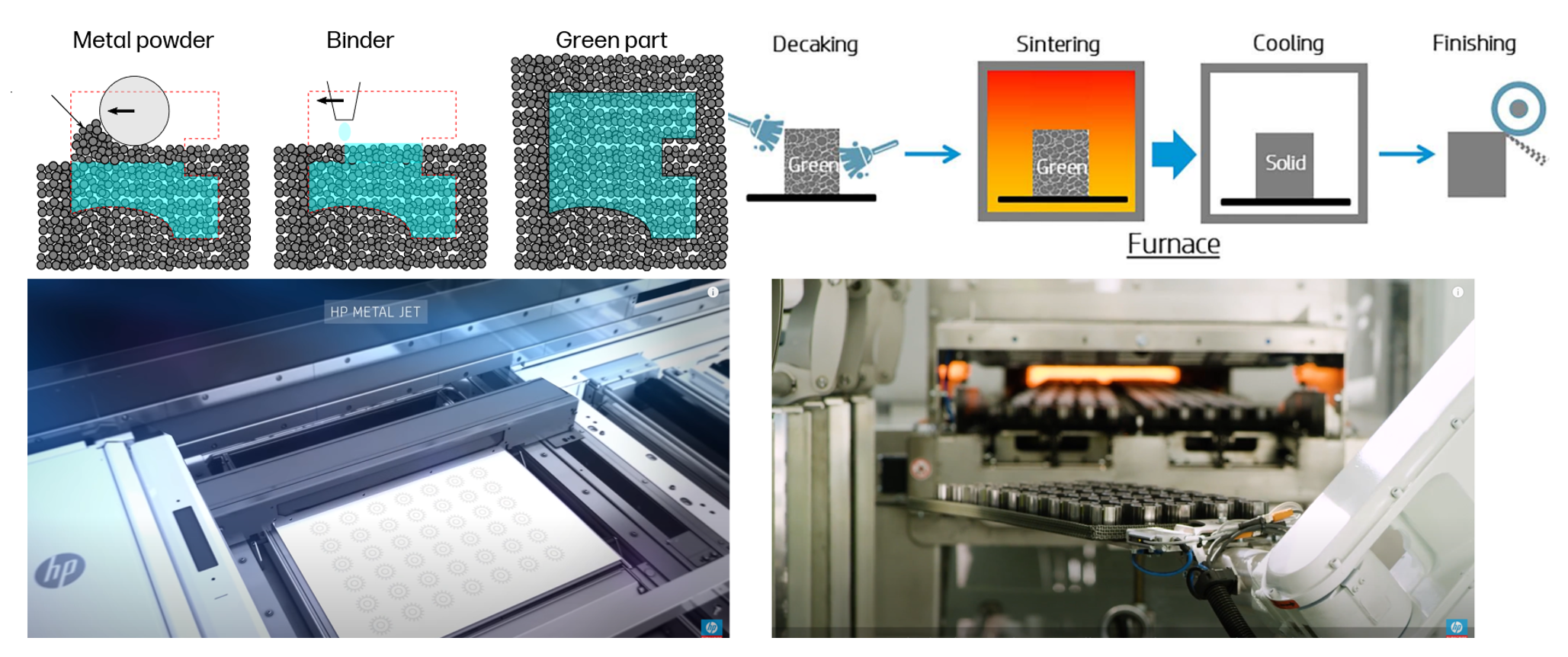}%
\caption{Metal Jet print stages to obtain the final part. }
\label{fig1}
\end{figure}

\begin{figure}
\centering
\includegraphics[width=0.5\textwidth]{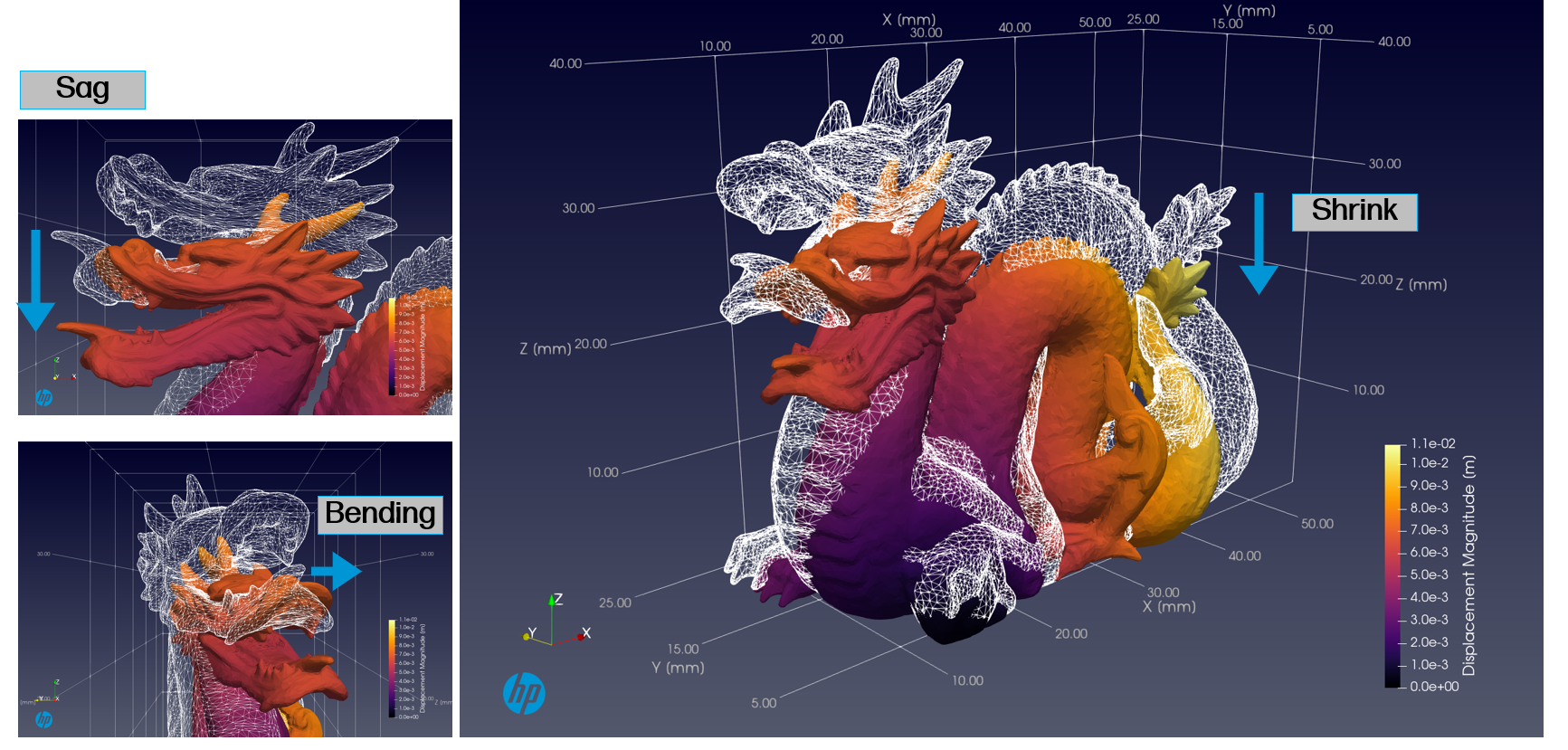}%
\caption{The Stanford dragon test model shows the isotropic part shrinkage, including gravitational sag, slump, and bending. }
\label{fig2}
\end{figure}

	Commercial software solutions for metal sintering deformation prediction exist, such as Abaqus and HP's organic solution for mental sintering simulation Virtual Foundry provides the domain science drive prediction based on the first principle of sintering physics. Virtual Foundry incorporates factors including thermal profile, yield curve, part porosity, and surface friction, which can simulate a part's deformation due to shrinkage, sagging effects, etc. As a physics-based simulation engine, Virtual Foundry today requires very small timesteps in simulation to manage the nonlinearity that arises from the sintering physics. Simulating a part's transient, dynamic sintering process can take from tens of minutes to several hours, depending on part size.  Many additional use cases can be opened if we can substantially speed up Virtual Foundry.

In this paper, as a component of HP’s Digital Twin effort, we propose the Virtual Foundry Graphnet led by HP Labs to predict the part deformation, which can speed up the deformation simulation substantially at the voxel level. Running a well-trained Metal sintering inferencing engine only takes a range of seconds to obtain the final sintering deformation value. The tested accuracy on example complex geometry achieves 0.7um mean deviation for a 63mm testing part, for a single sintering step (equivalent to 8.3 minutes physical sintering time), and a 0.3mm mean deviation for the complete sintering cycle (~4 hrs physical sintering).  This disclosed technology is our path to create a stand-alone deep-learning framework that can make fast (near real-time) and accurate end-¬¬¬to-end predictions of the sintering part deformation.

\section{Related Work}

\subsection{Graph neural networks }

Graph data structures are widely used in complex physical simulations, such as weather systems, fluid dynamics, and so on. The objects processed by these simulation tasks are often distributed in non-Euclidean spaces, and there are complex physical interactions between these objects. Such information is difficult to process with conventional convolutional neural networks or recurrent neural networks. Graph neural networks (GNNs) can represent such data on graph space well and can simulate the interaction between objects through message passing and aggregation. Due to its powerful data representation and inference capabilities, GNN has been widely applied in the field of physical simulation. Graph neural networks leverage deep leaning methods to graph-structured data,  it mainly has the following  branches: 

\begin{itemize}
  \item Recurrent graph neural networks,  aim to learn node representations with recurrent neural architectures\cite{Wu_2021}.
  \item Spectral convolution networks, use the spectrum of graph Laplacian to perform convolutions in the spectral domain. They have strong theoretical support\cite{bruna2014spectral}\cite{kipf2017semisupervised}.
  \item Spatial convolution networks, use message passing to aggregate local information from neighborhoods. They do not have much theoretical support, but they have simple architecture and have produced state of art results on some graph tasks\cite{atwood2016diffusionconvolutional}\cite{veličković2018graph}.
\end{itemize}

We use spatial convolution networks based on predecessors’ work\cite{sanchezgonzalez2020learning}.

\subsection{Digital Twin }
A Digital Twin is an integrated multi-physics, multiscale, probabilistic simulation, as a virtual replica of a real physical asset or system. It is a computational model that collects online data and information, based on physical theories, produces accurate and synchronized simulations, or forecasts the future of the physical counterpart\cite{DBLP:journals/corr/abs-1911-01276}\cite{inproceedings-glaessgen}\cite{MASHALY2021299}.  A Digital Twin system, through the rapid (real-time) collection of online data and information analytics, enables engagement with the physical environment, updating changes (performance, maintenance) of the physical system \cite{inproceedings-liu}, making real-time recommendations and autonomous decisions to facilitate the operation in the physical world. Digital Twin enables the Industrial 4.0 revolution, extending to industries including manufacturing, healthcare, and smart cities. For example, GE developed the “Predix” platform for creating Digital Twins to run data analysis and monitoring, and “imodel.js” created by Bentley Systems is another example of a platform for building Digital Twins\cite{systems7010007}. 

\subsection{Sintering physics  }
The manufacturing process of a particular material from powder consists of the consolidation step (i.e. injection molding, extrusion, slip, isostatic pressing, etc.)  and the sintering step that produces the final densified product. The sintering process is the powder metallurgy process during the thermal treatment which contacting particles are bonded to form the objects of the required mechanical properties. Sintering occurs primarily at the atomistic level, during which the rheological response of the powder body as a homogeneous, porous continuum can be described by the governing constitutive laws\cite{article-blaine}.  The sintering kinetics of particulate bodies accounts for the impact from all properties: particles themselves, the local interactions with each other, macroscopic factors (externally applied forces, kinematic constrictions, i.e., adhesion of the part and furnace), inhomogeneity of the initial density distribution before sintering, etc. adient penalty. 

	Sintering simulation software tools such as the Finite element code Abaqus use the constitutive laws of sintering based on the continuum model. HP’s in-house software Virtual Foundry (Digital Sintering) also considers the bodies as homogeneous, porous continuums and simulates plastic deformation at elevated temperatures governed by diffusional theories and geometrical models of grains, grain boundaries, and pores. As a physics-based simulation engine, Virtual Foundry today requires very small timesteps in sintering physics simulation, leading to minutes to several hours for simulating a part's transient, dynamic sintering process depending on part size, limiting the user applications.  
 
\subsection{Physics informed network  }

	With the recent advances of deep neural networks in various applications, including computer vision and natural language processing, \cite{article-nlp} applying neural networks to scientific problems has been investigated. For example, Alphafold\cite{article-alphafold} predicts protein structures successfully, and FourCastNet\cite{pathak2022fourcastnet} forecasts weather and the traditional numerical weather simulation. However, the expressive power of neural networks is limited by data size, it is only easily applicable to some scientific problems where data acquisition is not highly limited or expensive. A physics-informed neural network \cite{cuomo2022scientific} is designed to address such problems by incorporating prior knowledge of physics. PINN proposed to solve the PDE system by adding extra constraints called physics-informed loss on neural networks. DeepONet uses two networks, TrunkNet and BranchNet; BranchNet is used for encoding and processing the data, and TrunkNet is used for physics-informing the model at any query point. Fourier neural operator maps from the physical domain into the frequency domain then apply convolutions on the Fourier domain, thus building dimension-free data-driven neural physics solutions. However, PINN, DeepONet, and Fourier Neural Operator are based on Euclidean grid structure. Therefore, it is not easily scalable to scientific problems defined on non-Euclidean domains. Graph Neural Networks and its variants \cite{xu2019powerful} have been proposed to address the issue by applying convolutions on the graph domain, such that one can solve the PDE system on non-Euclidean domains.

 	For complex physical simulation tasks, such as weather systems and fluid dynamics, objects processed are often distributed in non-Euclidean spaces and contain complex physical interactions. Graph neural networks (GNNs) can represent such data on graph space and can simulate object interaction better than conventional convolutional or recurrent neural networks. 
 
\section{The proposed method}
We proposed a graph network-based simulator to build out a quantitative model that can learn the entire physics simulation process and deliver much faster prediction. The deep neural network could:

\begin{itemize}
  \item {Directly infer the sintered object deformation state (displacement vectors at the voxel level, and other physical fields associated with voxels, for example, porosity) at time $T(t=k)$, when given the state of the object deformation at time $T(t=0)$ to $T(t=k-1)$, $k > 1$.}
  \item Rollout to infer the sintered object deformation state after $T (t=k)$,  i.e., $T(t=k+1), T(t=k+2), …  T(t=k+n), n>1$, with given the state of the object deformation at $T(t=0)$ to $T(t=k-1)$. In our tested setting, $k$ can be larger than 50, and possibly be even larger with optimized model training to cover the entire sintering simulation. 
  \item Such inferencing using sintering simulation state at $T(t=0)$ to $T(t=k-1)$ as input predicts the sintering simulation state at $T(t=k)$, is done near real-time, for example, less than a second. It predicts the entire rollout simulation states, i.e. $T(t=k+1), T(t=k+2), …  T(t=k+n)$ in less than or around 10 seconds (depending on the input geometry size). This is much faster than obtaining each sintering state through FEA simulation – in which case, many timesteps will be required, and each timestep may take minutes to complete. Using our proposed inferencing model to replace physical simulation achieves a much faster speed-up for Virtual Foundry. 
\end{itemize}

\subsection{Data preparation }

\begin{figure}
\centering
\includegraphics[width=0.5\textwidth]{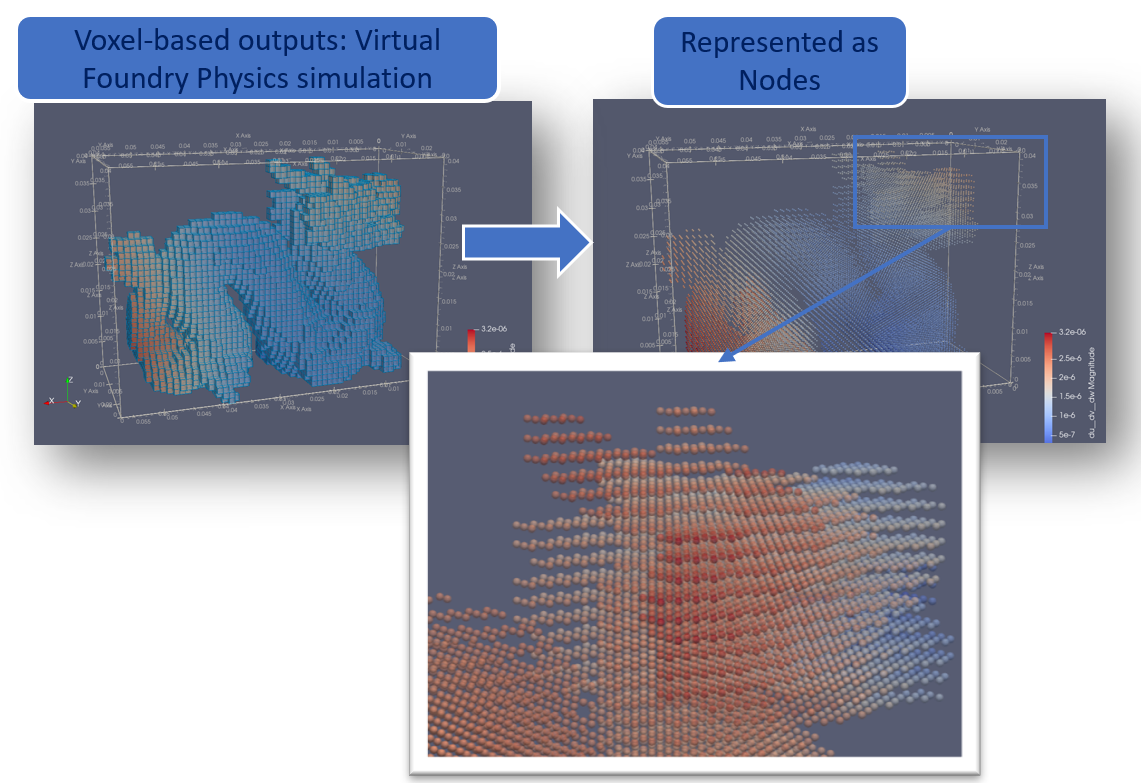}%
\caption{ Data process pipeline to obtain the sintering-time graph data for training and inferencing. The sample dragon build is voxelated, the color gradient shows each voxel’s deformation scale; we preprocess the voxel-based simulation output (left) to form “nodes” of the graph (right), representing the concept of “metal particles”. }
\label{fig3}
\end{figure}

\begin{figure}
\centering
\includegraphics[width=0.5\textwidth]{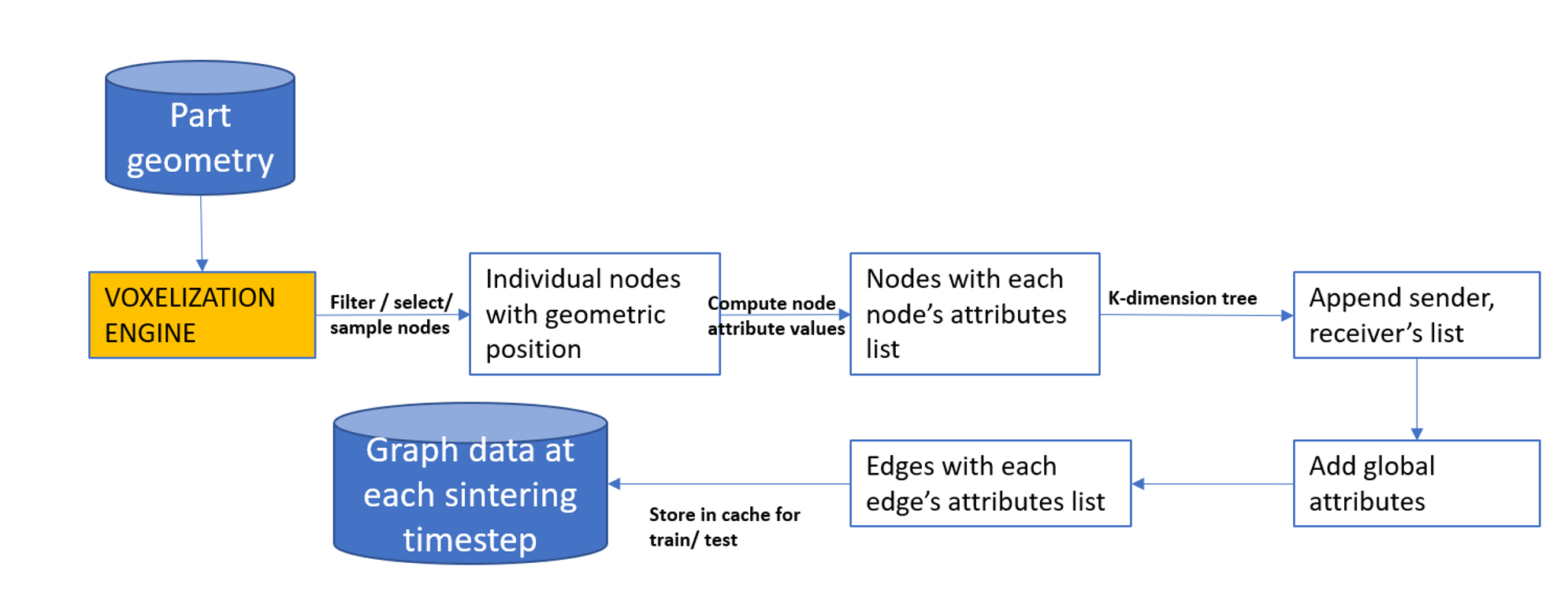}%
\caption{Data process pipeline to obtain the sintering-time graph data for training and inferencing. The preprocessing steps are to convert the voxel-based simulation data into a graph-based data structure for each sintering step }
\label{fig4}
\end{figure}

\begin{enumerate}
  \item Virtual Foundry transfers the origin design file (in the form of STL, OBJ, 3MF, etc.) of parts into voxel data through a voxelization process and outputs voxelated format representing part 
  \item Based on the voxelated data, we compute vertices and edges of voxel data to build a graph $G(\mathcal{V}, \mathcal{E}, \mathcal{U})$.  $G$ represents a graph structure model at a specific sintering timestep. $\mathcal{V}$ represents the nodes and $\mathcal{E}$ represents the edges between two nodes, the connection of two nodes through the edge represents the interactions between the two nodes.  \\
        Each node in the graph has a list of attributes, as shown in Fig. \ref{fig4}, the attribute list indicates the features of this node. For example, the node attribute list of node $i$ is in the form of  $[\mathbf{s}_{k-2}, \mathbf{s}_{k-1}, \mathbf{s}_{k}, {T}_{k}]$, where $\mathbf{s}_{k-2}, \mathbf{s}_{k-1}, \mathbf{s}_{k}$ is node $i$’s previous three timestep’s velocity, each $\mathbf{s}_{k}$ is a three-dimensional vector describing velocity in $XYZ$-dimensions. \\
        Initial $n$ (the equation shows the case when $n$=3, we tested with various $n$ values such as $n$=2, $n$=6) timestep’s velocity vectors are added as node attributes, representing the moving speed of each node. Sintering physics has an intrinsic memory effect, that is, the current state has a strong dependency on the previous state's history.  Other physics-inspired parameters can be added as node attributes, such as node velocity, and accelerations vector of each timestep. \\
        Boundary constraints were also implemented to distinguish nodes that can move with all degrees of freedom from the nodes that fall on the contacting surface. 
        
  \item To define the neighbors of each node, we build directed edges among nodes using a k-dimensional tree and find neighbors of each node within a radius $r$. Set node $\mathcal{V}_{i}$ as the \textbf{“sender”} of a directed edge, nodes within its radius r range are included as the \textbf{“receivers”} of the directed edge of this node. We define radius $r$ to be 1.2 times voxel size so that each node has 6 neighbors as receivers. Each edge in the graph also has a list of attributes, including the relative distance of the connected nodes. 
  \item Optionally, we implement an additional component $\mathcal{U}$, representing the global factors of the entire graph. For example, we use sintering temperature at different times as a component $\mathcal{U}$.   
  \item At different sintering timesteps, the graph structure changes, such as the number of edge connections among nodes (indicating the dynamics of node interactions), nodes and edges’ attribute value (indicating the change of displacement value at each node, change of stress), etc.  
\end{enumerate}

\subsection{	Model architecture and training }


\begin{figure}
\centering
\includegraphics[width=0.5\textwidth]{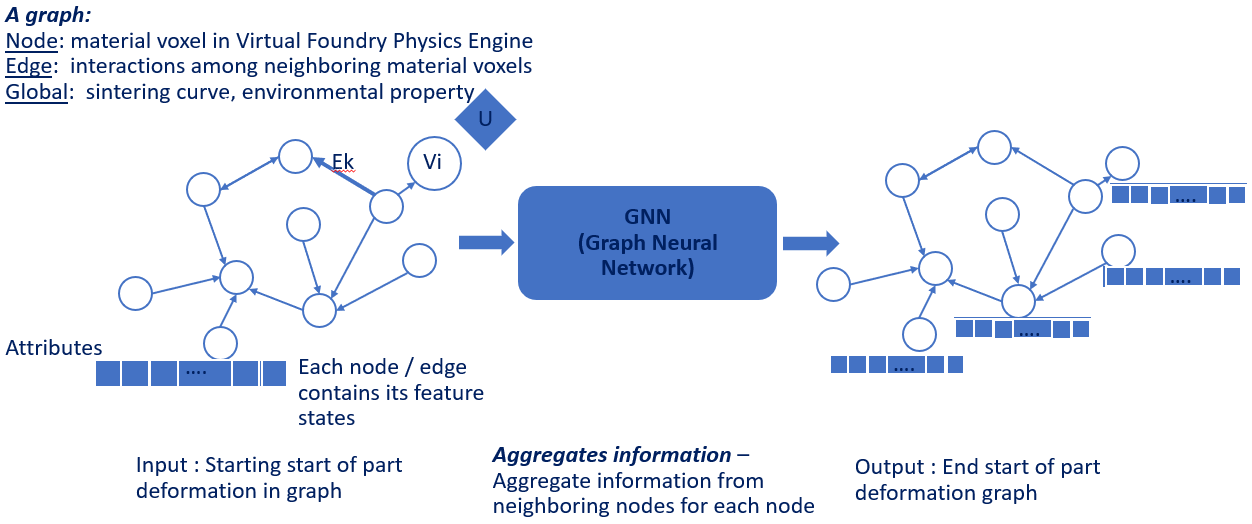}%
\caption{Simulation prediction on the processed graph data with graph neural network architecture }
\label{fig5}
\end{figure}

The learned simulator consists of three stages: encoder, processor, and decoder. The encoder converts Virtual Foundry data to latent graph data $G(\mathcal{V}, \mathcal{E})$. here, $\mathcal{V}, \mathcal{E}$ are sets of vertices and edges, respectively. The processor computes interactions among nodes, aggregates information, and outputs learned latent features. The processor has an interaction networks module.  Firstly, it uses the previous step’s features to update edge features (edge attributes list). Then it uses the previous feature set and the updated edge features to update node features. Finally, it uses previous features, edge features, and node features to update global features.  

We use multilayer perceptron (MLP) for each update function. The weights and parameters of each MLP are learned during training. We perform multiple rounds of this calculation, each round is referred to as one “message passing” step, representing the interaction among nodes through edges. The number of messages passing rounds can vary depending on the training setting. In our case, we set it equal to 10. The more message-passing rounds, the more accurate the prediction in general, since more message-passing rounds represent the larger node interaction scope. We chose the optimal trade-off setting between accuracy and training runtime.  

The decoder extracts displacement data from learned latent features and sends it to Virtual Foundry. For more details, please refer to the section on integrating DL inside Virtual Foundry. 

\subsection{Objective functions }

Our loss function design is as below, at time point $k$, features with length $n$ as the input of the graph neural networks $f$ and accelerations with length $l$ as the output of the graph neural networks $f$. Instead of predicting one-step acceleration after time point $k$, we predict and constrain multiple steps’ node accelerations. That is, the graph networks can directly or recurrently output the elements of $Y$. 

\begin{align}
    \mathbf{Y} &=  [ \mathbf{a}_{k+1}, \mathbf{a}_{k+2}, \mathbf{a}_{k+3}, ..., \mathbf{a}_{k+l} ] \\
    \mathbf{X} &= [\mathbf{s}_{k-n}, ..., \mathbf{s}_{k-2}, \mathbf{s}_{k-1}, \mathbf{s}_{k}, \textbf{g}]  \\
     \mathbf{Y} &=  {f_{\theta}(\textbf{X})}
\end{align}

Metal sintering is a complicated physics process. It’s difficult to predict long-term deformation without a large dataset. Instead of predicting one step after time point $k$, our method predicts $l$ consecutive small steps, $l$ can be 1 step or multiple steps. Multiple steps are chosen to prevent the graph networks only learning short-term dynamics of sintering. This can potentially alleviate the overfitting of training and reduce the prediction time of rollout. We use mean square error $L(Y|X)$ to optimize the graph networks and add a discount factor in the loss, where $\textbf{a}$’ denotes the predicted values, and $\textbf{a}$ denotes the ground truth values:  

\begin{equation*}
    \begin{aligned}
            L(Y|X) = (\mathbf{a}^{'}_{t_{k+1}} - \mathbf{a}_{t_{k+1}})^2 + 
   \mathcal{\gamma} (\mathbf{a}^{'}_{t_{k+2}} - \mathbf{a}_{t_{k+2}})^2 + \\
   \mathcal{\gamma}^2 (\mathbf{a}^{'}_{t_{k+3}} - \mathbf{a}_{t_{k+3}})^2 +  
   ... + 
   \mathcal{\gamma}^{l-1} (\mathbf{a}^{'}_{t_{k+l}} - \mathbf{a}_{t_{k+l}})^2 
    \end{aligned}
\end{equation*}

Additionally, apart from differentiating the node types and adding the type of vector as a node attribute for the model to learn differentiating during training, we can also constrain the “Fixed node” and “Slip nodes” moving explicitly by setting their velocity or acceleration equal to 0 in the model loss function (training objective function). With this loss function implementation, the model learns with explicit instruction that different node type has 0 acceleration in stated dimensions – not moving in the stated dimensions.

\begin{figure}
\centering
\includegraphics[width=0.5\textwidth]{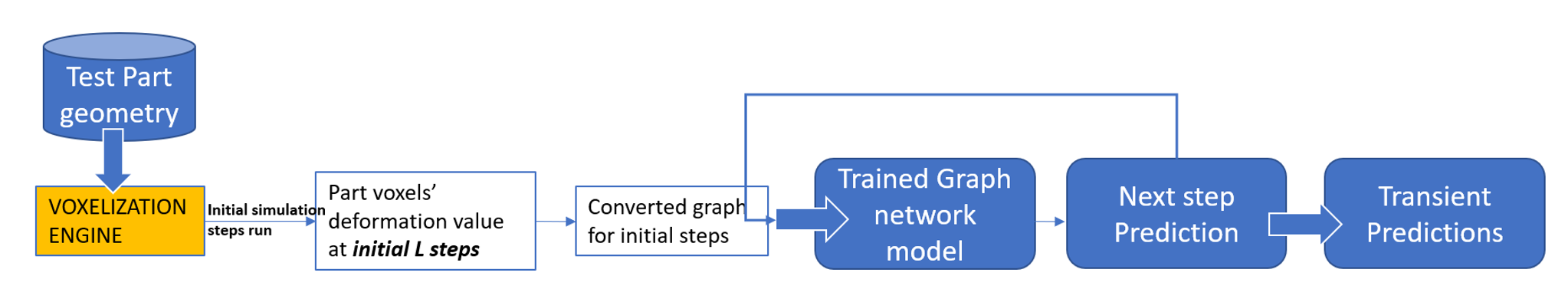}%
\caption{Inference pipeline to iteratively predict the next sintering state.}
\label{fig6}
\end{figure}

\subsection{Inferencing}
Like the data preparation in the model training process, the converted graph for the initial steps serves as the input of trained graph networks. Corresponding to the training set, that is, the model takes one data sample from $t=0$ to $t=n$ (the initial $n$ steps, where $n$ is equal to the model training parameter setting) as input, $G(\mathcal{V}, \mathcal{E}, \mathcal{U})$
at $t=n+1$  is to be predicted. Firstly, the trained graph network makes the next timestep prediction. 
Then, the output of trained graph networks $G(\mathcal{V}, \mathcal{E}, \mathcal{U})$ at $t = n +1$  is recursively fed as input and builds up the following timestep’s graph input (from $t = 1$ to $t = n + 1$) to the same trained graph networks to predict the next timestep’s $(t = n + 2)$  graph features. This process is repeated until the entire sintering process prediction is completed.

\section{Experimental results}

\begin{table*}[!t]
\centering
\begin{tabular}[t!]{c c c c c c c c} 
 \hline
{
\textbf{Experiment}} & \textbf{Model}  & \textbf{$T$}  & \textbf{Edge}  & 
 \textbf{Node-feature}  & \textbf{Anchor plane} & \textbf{1-step MSE} & \textbf{Rollout MSE} \\
 
\# & \textbf{version} & \textbf{profile} & \textbf{dropout} & \textbf{normalize} & \textbf{\& anchor point $l$} &  ($\mathbf{e}^{10-9}$) & ($\mathbf{e}^{10-2}$)\\ [0.5ex] 
 \hline\hline
 
 \multirow{2}{*}{1} & GNS & \multirow{2}{*}{Yes}  &\multirow{2}{*}{-}  & \multirow{2}{*}{-}   & \multirow{2}{*}{-}  & \multirow{2}{*}{49.7 / 95.7} & \multirow{2}{*}{52.1 / 43.8}\\ 
   &  +  Temperature($T$) & & & & & \\
 \hline
 
 \multirow{2}{*}{2}& GNS+ $T$  & \multirow{2}{*}{Yes}  & \multirow{2}{*}{Yes} & \multirow{2}{*}{-}  & \multirow{2}{*}{-}  & \multirow{2}{*}{5.9 / 1.6} & \multirow{2}{*}{0.3 / 0.1}\\
   &  +Edge dropout& & & & & \\
 \hline
 \multirow{2}{*}{3} & M\#2 + Node  & \multirow{2}{*}{Yes}  & \multirow{2}{*}{Yes} & \multirow{2}{*}{Yes}  &\multirow{2}{*}{-} & \multirow{2}{*}{1.9 / 0.2} & \multirow{2}{*}{0.1 / 0.5}\\
    & feature normalize & & & & & \\

 \hline
 \multirow{2}{*}{\textbf{4}} & \textbf{M\#3 +Anchoring}& \multirow{2}{*}{Yes}  & \multirow{2}{*}{Yes} & \multirow{2}{*}{Yes} & \multirow{2}{*}{Yes} & \multirow{2}{*}{\textbf{1.5 / 0.6}} & \multirow{2}{*}{\textbf{0.1/0.1}} \\
 & \textbf{Loss} & & & & & \\
 \\
 [1ex] 
 \hline
\end{tabular}
\caption{Model version accuracy comparison.}
\end{table*}
\label{Table. 1. }

NVIDIA GPUs are used to train/test the model. We implemented the model with Pytorch with parallel training and mixed precision.  Our code is integrated into the NVIDIA modulus platform as an application example of the Physcis-ML models.

\subsection{Model optimization}

We trained the proposed model on a limited number of geometries (7 different geometries at mesh size equal to 1000 microns are included in the training data to produce the Table 1 testing results). We tested both the “1-step accuracy” (the prediction accuracy of the very next timestep) and “Rollout accuracy” (the prediction accuracy after the entire sintering process) on different geometries. Table 1 shows the version optimizations based on our baseline graph networks model. The improved model versions and their improved measured accuracy are listed. We tested model architecture and hyperparameter variations, including adding sintering temperature as the global feature, implementing edge dropout (with random edge dropout ratio set to be 60\%, the randomly selected edges are updated for every training epoch) during model training, normalizing the input node features, including node velocities and accelerations and adding additional anchoring point losses. Model version \#4 shows the best prediction accuracy in both the next step and rollout predictions. We adopted the best model architecture version for later tests and report the test accuracy in the next section.

\subsection{Test geometry accuracy }

	We demonstrate the model testing accuracy in this paper on four different test geometries and different voxelization resolutions (voxel size equal to 1000 microns, 500 microns); the details of the test parts, including the number of voxels and the number of nodes for each test geometry after data preprocessing are as shown in Table 2. Figure 6 shows the simulation complete state result of the part deformation compared to the sintering start time (with no deformation). The color bar shows the displacement magnitude in millimeters.   

 \begin{figure}
\centering
\includegraphics[width=0.5\textwidth]{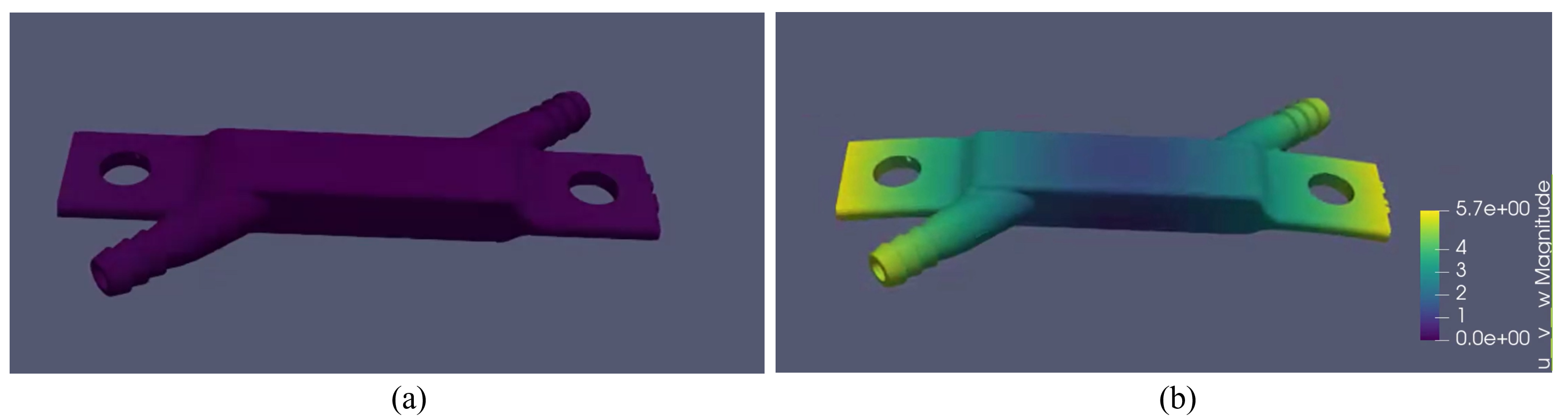}%
\caption{(a) Sample geometry Busbar at sintering start time shows no deformation across the part. (b) Part deformation at the sintering finish state. The simulation was generated with a mesh size equal to 500 microns.}
\label{fig7}
\end{figure}

\begin{figure}
\centering
\includegraphics[width=0.5\textwidth]{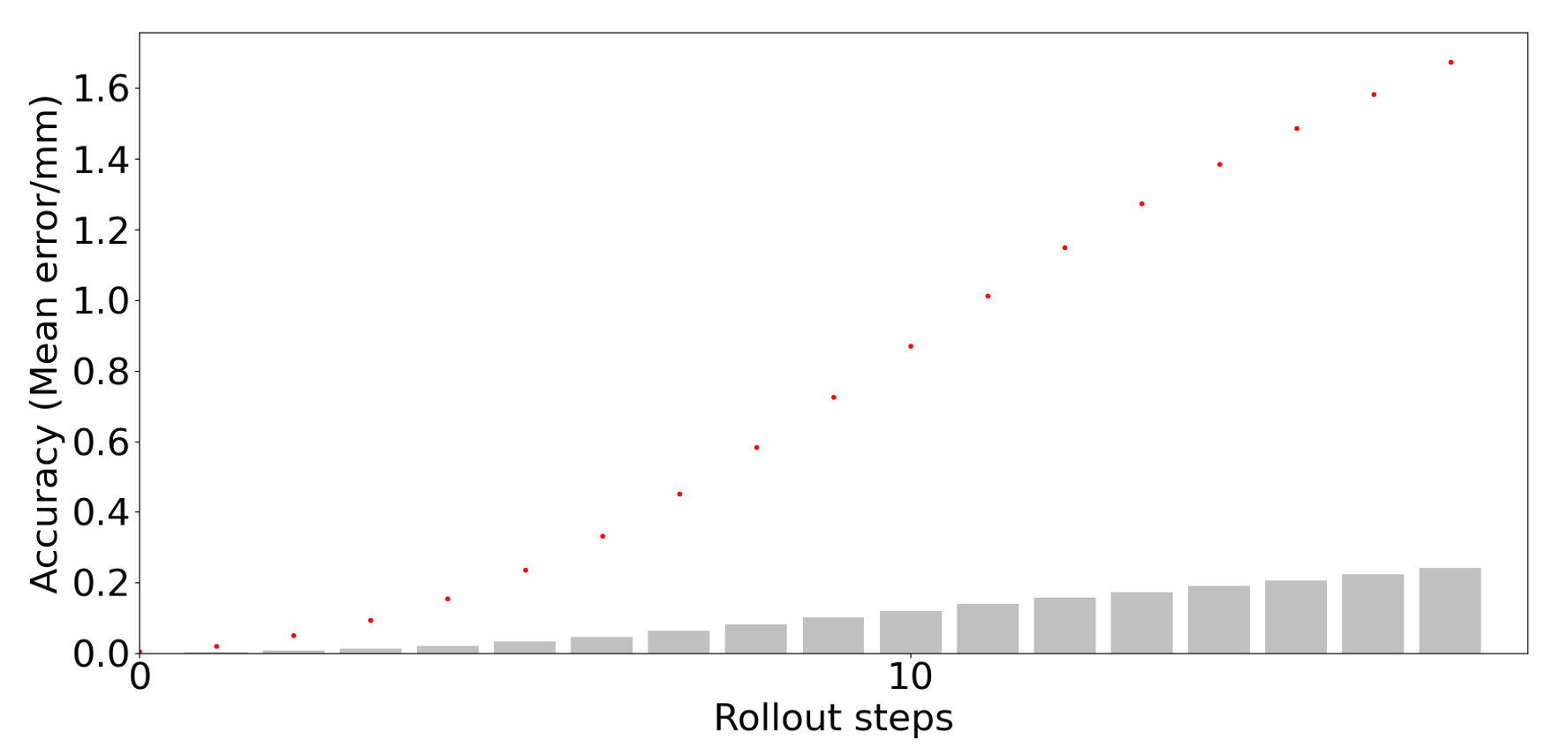}%
\\
\includegraphics[width=0.5\textwidth]{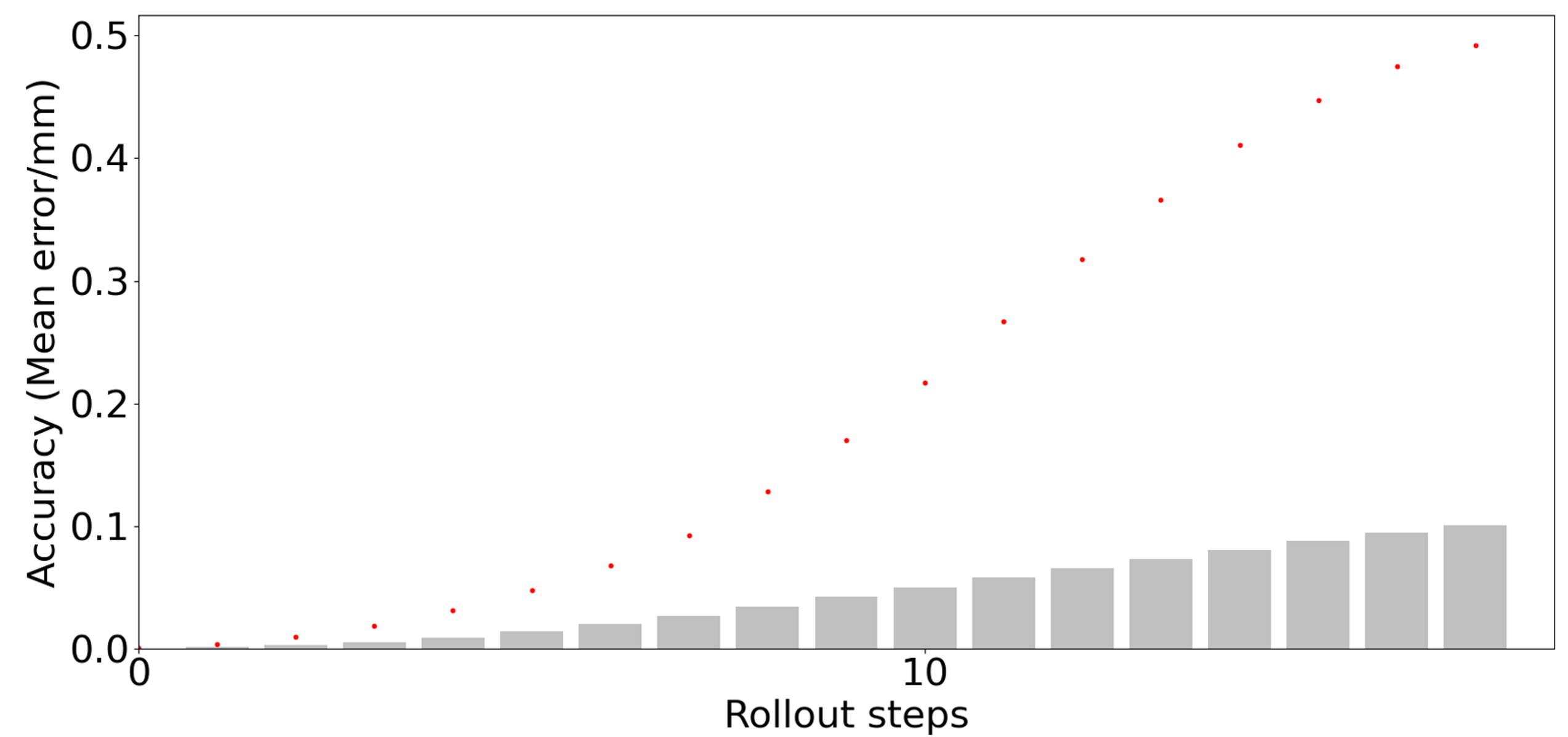}%
\caption{Rollout prediction accuracy on two tested geometries (a) USB and (b) Busbar. The bar plot shows the mean nodal prediction accuracy, and the dot plot (in red) shows the maximum nodal prediction accuracy.}
\label{fig8}
\end{figure}

\begin{figure}
\centering
\includegraphics[width=0.5\textwidth]{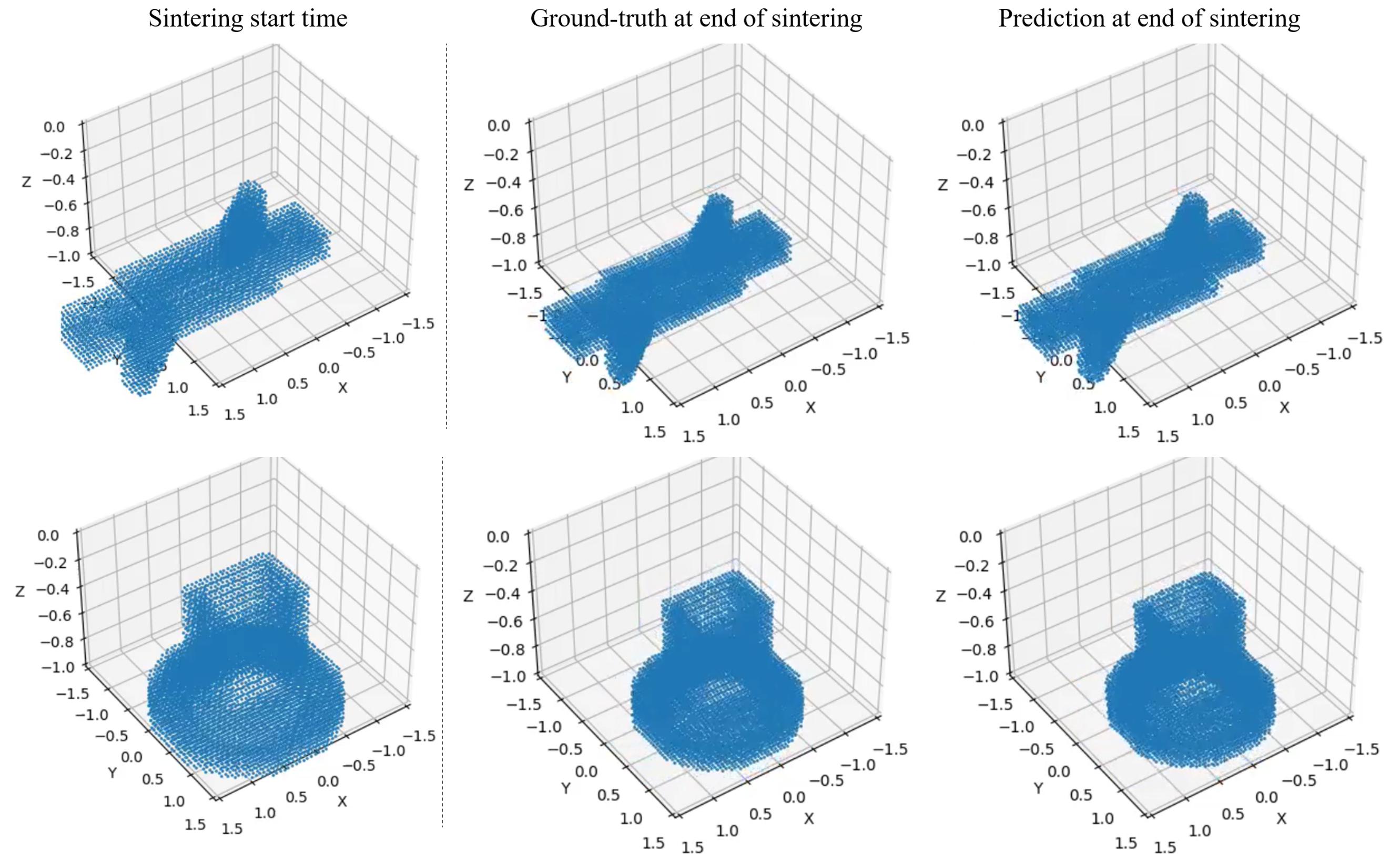}%
\caption{Rollout prediction accuracy on two tested geometries (a) USB and (b) Busbar. The bar plot shows the mean nodal prediction accuracy, and the dot plot (in red) shows the maximum nodal prediction accuracy.}
\label{fig9}
\end{figure}

	We demonstrate the rollout prediction accuracy on two tested parts in Fig.\ref{fig8}. The bar plot shows the mean nodal prediction accuracy (measured as a mean error in millimeters) until the end of the end-of-sintering cycle, and the dot plot in red shows the maximum nodal prediction accuracy (measured as a mean error in millimeter). The mean nodal predicted difference for the first test part (the USB) is below 0.3mm, and the max nodal predicted difference is around 1.2mm. For the second test part (the Busbar), the mean nodal predicted difference is around 0.2mm, and the max nodal predicted difference is around 1mm at the end of the sintering simulation process. Figure \ref{fig9} illustrates the nodal-level deformation sampling at a mesh size equal to 1000 microns at the sintering start time (with no deformation) and sintering completion state, comparing the simulation ground truth value and the model predicted value.

\begin{table*}[!t]
\centering
\begin{tabular}[t!]{c c c c } 
 \hline

 \textbf{Test geometry }  & \textbf{Mesh size = 1000} & \textbf{Mesh size = 500} & \textbf{Size} \\
 
\textbf{(voxels / nodes count) } & \textbf{($10^3$)} &  \textbf{($10^3$)} & \textbf{(Diagonal / mm)}\\ [0.5ex] 
 \hline\hline
 
  \multirow{2}{*}{Extrusion Screw}   & \multirow{2}{*}{0.8 / 6.8}  & \multirow{2}{*}{4.8 / 38.7} & \multirow{2}{*}{19}\\ 
   & & & \\
 \hline
 
  \multirow{2}{*}{USB}  & \multirow{2}{*}{5.5 / 44.3}  & \multirow{2}{*}{35.3 / 282.0} & \multirow{2}{*}{59}\\
   & & & \\
 \hline

   \multirow{2}{*}{Pushing Grip}  & \multirow{2}{*}{10.6 / 84.4}  & \multirow{2}{*}{79.9 / 639.4} & \multirow{2}{*}{87}\\
   & & & \\
 \hline

    \multirow{2}{*}{Busbar}  & \multirow{2}{*}{3.8 / 30.6}  & \multirow{2}{*}{20.5 / 11.8} & \multirow{2}{*}{63}\\
   & & & \\
 \hline
 
\end{tabular}
\caption{Test part geometry size}
\end{table*}
\label{Table. 2. }

\begin{figure}
\centering
\includegraphics[width=0.5\textwidth]{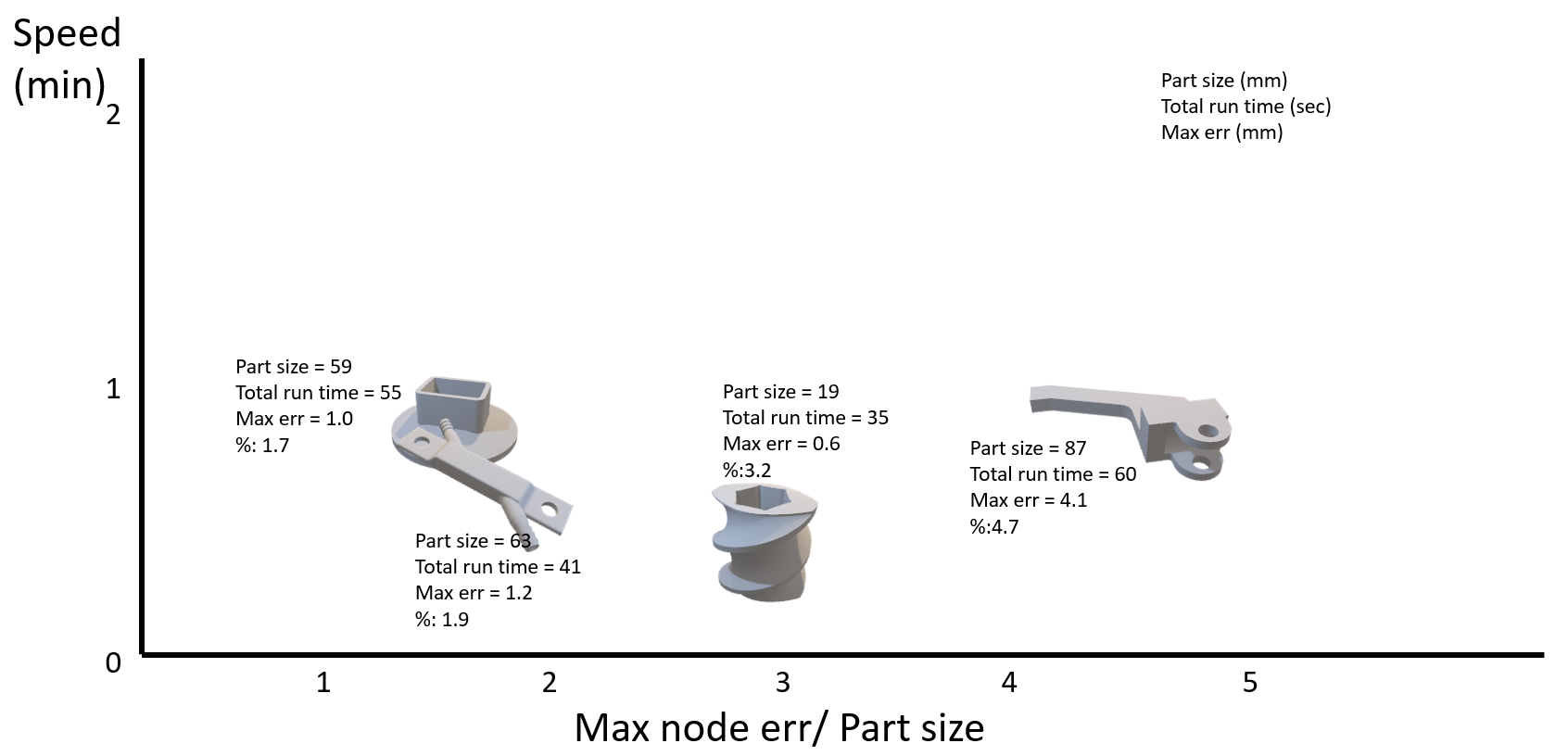}%
\caption{Max node error (in mm, and \% to part size), inference runtime (to obtain the final deformation prediction) v.s. part original size, on the four tested geometries (left to right: USB, Busbar, Extrusion screw, Pushing grip)}
\label{fig10}
\end{figure}

 	We demonstrate the test result on all four geometries at a mesh size equal to 1000 microns, as shown in Fig. \ref{fig10}: the USB with the part size of 59mm (measured the diagonal length), Busbar (part size of 63mm), Extrusion screw (part size 19mm), and Pushing grip (part size 87mm). The inference model was able to scale well with a small accuracy tradeoff at a smaller mesh size (we tested sample parts at 500 microns and 300 micros) in our tests. Fig. \ref{fig10}. Shows the inference runtime to obtain the final deformation prediction for all four parts within or around one minute, the runtime is computed with the input generation time from Virtual Foundry and the trained network inferencing time. Two tested parts have a maximum node error of less than 2\% (compared to the part size). The Pushing grip shows the most significant maximum nodal error, yet it is still within 5\%. This inference run simulated the actual physical sintering time of around 3 hours.

\section{CONCLUSION AND FUTURE WORK}
With the model prediction accuracy and fast inference speed. This work as a component of HP’s Digital Twin effort, Virtual Foundry Graphnet led by HP Labs aims to apply Physics-Informed Neural Networks (PINN) to greatly accelerate the computation that predicts the metal powder material phase transition. It has achieved orders of magnitude speed-up compared to physics simulation software while preserving reasonable accuracy.  Furthermore, Virtual Foundry Graphnet has demonstrated an outstanding path-forward to scaling for diverse parts of arbitrary geometrical complexity, and scaling for different process parameter configurations. 

We at HP Labs recognize the significant role a growing open-source community can play in accelerating PINN’s development and expand its applications; Nvidia’s Modulus provides an outstanding platform to host such a PINN open-source community.  With open-sourcing Virtual Foundry Graphnet at the Modulus platform, HP Labs are joining the Modulus open-source community. 

\section*{Acknowledgment}
This material is based upon work supported by the Air Force Office of Scientific Research under award 
number FA9550-23-1-0739.
Any opinions, findings, and conclusions or recommendations expressed in this material are those of the 
author(s) and do not necessarily reflect the views of the United States Air Force.

\bibliographystyle{IEEEtran} 
\bibliography{test.bib}

\begin{IEEEbiography}[{\includegraphics[width=1.0in,height=1.2 in,clip,keepaspectratio]{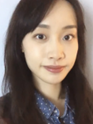}}]
{Rachel (Lei) Chen} is a Machine Learning Research Engineer at HP Inc. She received her M.S. degree in Electrical and Computer Engineering from Duke University in 2019 and her B.S. degree in Electrical Engineering from Korea Advanced Institute of Science and Technology (KAIST) in 2017. She is now devoted to research in applying cutting-edge deep learning algorithms in Additive Manufacturing quality control, acceleration, and Digital Twins. 
 (lei.chen1@hp.com)
\end{IEEEbiography}

\begin{IEEEbiography}[{\includegraphics[width=1.0in,height=1.2 in,clip,keepaspectratio]{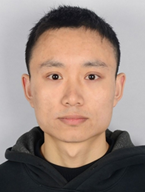}}]
{Chuang Gan} received his B.S. degree from Nanjing University of Aeronautics and Astronautics, China, in 2014 and his M.S. degree from Zhejiang University, China, in 2017. Since 2017, he has been a software engineer and researcher at HP Inc. His research interests are in deep learning and computer vision.
(chuang.gan@hp.com). 
\end{IEEEbiography}

\begin{IEEEbiography}[{\includegraphics[width=1.0in,height=1.2 in,clip,keepaspectratio]{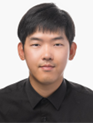}}]
{Juheon Lee} received his PhD degree from the University of Cambridge, UK, in 2016. Since 2019, he has been a research scientist at HP Inc. His research interests are in geometric deep learning physics-informed neural networks and neural tangent kernel theory.  
(juheon.lee@hp.com). 

\end{IEEEbiography}

\begin{IEEEbiography}[{\includegraphics[width=1.0in,height=1.2 in,clip,keepaspectratio]{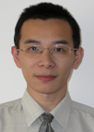}}]
{Zijiang Yang} received his B.S. degree from Zhejiang University, China, in 1997 and his Ph.D. degree from Zhejiang University, China, in 2003, respectively. From 2003 to 2008, he was a senior software specialist at Nokia, China. Since 2008, he has been a software expert and master technologist at HP Inc., China. His research interests are in deep learning and IoT. 
(zijiang.yang2@hp.com).

\end{IEEEbiography}
\begin{IEEEbiography}[{\includegraphics[width=1.0in,height=1.2 in,clip,keepaspectratio]{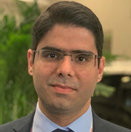}}]
{Mohammad Amin Nabia} is a senior software engineer, at AI-HPC at NVIDIA. Mohammad is one of the core developers of NVIDIA Modulus, an AI framework for physics-ML models. Mohammad received his Ph.D. from the University of Illinois at Urbana-Champaign, with research focused on artificial intelligence for scientific computing. Mohammad's background is in numerical simulation, artificial intelligence, and uncertainty quantification.
(mnabian@nvidia.com).

\end{IEEEbiography}

\begin{IEEEbiography}[{\includegraphics[width=1.0in,height=1.2 in,clip,keepaspectratio]{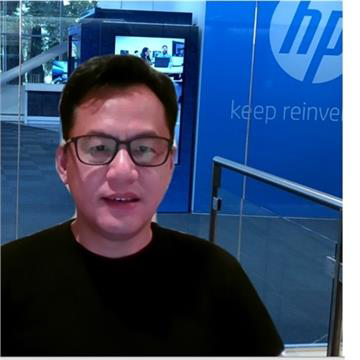}}]
{Jun Zeng} is a Distinguished Technologist at HP Inc. and a principal investigator and research manager leading software research in 3D Printing and Digital Manufacturing for the HP 3D Printing Software group. His publications include a co-edited book on the Computer-aided Design of microfluidics and biochips, a co-authored book on production management of digital printing factories, 50+ peer-reviewed papers, and 42 granted patents and more pending. His academic training includes advanced degrees in mechanical engineering (PhD) and computer science (M.S.), both from Johns Hopkins University. Jun is an ACM member and an IEEE senior member.
(jun.zeng@hp.com).

\end{IEEEbiography}

\end{document}